\documentclass[sigconf]{acmart}
\usepackage{amsmath}
\usepackage[ruled,vlined]{algorithm2e}
\usepackage{multirow}
\usepackage{enumitem}
\usepackage{color}

\AtBeginDocument{%
  \providecommand\BibTeX{{%
    \normalfont B\kern-0.5em{\scshape i\kern-0.25em b}\kern-0.8em\TeX}}}
\setcopyright{none}
\copyrightyear{2018}
\acmYear{2018}
\acmDOI{XXXXXXX.XXXXXXX}

\acmConference[Conference acronym 'XX]{Make sure to enter the correct
  conference title from your rights confirmation emai}{June 03--05,
  2018}{Woodstock, NY}
\acmISBN{978-1-4503-XXXX-X/18/06}

\begin{document}

\newcommand{\ML}[1]{{\color{teal}M.L.:[#1]}}

\title{KGExplainer: Towards Exploring Connected Subgraph Explanations for Knowledge Graph Completion}

\author{Tengfei Ma}
\email{tfma@hnu.edu.cn}
\affiliation{%
  \institution{College of Computer Science and Electronic Engineering, Hunan University}
  \country{China}
}

\author{Xiang Song}
\affiliation{%
  \institution{AWS AI Research and
Education}
  \country{USA}
  }
\email{xiangsx@amazon.com}

\author{Wen Tao}
\affiliation{%
  \institution{College of Computer Science and Electronic Engineering, Hunan University}
  \country{China}
}\email{taowen@hnu.edu.cn}

\author{Mufei Li}
\affiliation{%
 \institution{Georgia Institute of Technology}
 \country{USA}
 }\email{mufei.li@gatech.edu}


\author{Jiani Zhang}
\affiliation{%
  \institution{Amazon Web Services}
  \country{USA}
  }
\email{zhajiani@amazon.com}

\author{Xiaoqin Pan}
\affiliation{%
  \institution{College of Computer Science and Electronic Engineering, Hunan University}
  \country{China}
  }
\email{pxq123@hnu.edu.cn}

\author{Jianxin Lin}
\affiliation{%
  \institution{College of Computer Science and Electronic Engineering, Hunan University}
  \country{China}
  }
\email{linjianxin@hnu.edu.cn}

\author{Bosheng Song}
\affiliation{%
  \institution{College of Computer Science and Electronic Engineering, Hunan University}
  \country{China}
  }
\email{boshengsong@hnu.edu.cn}

\author{Xiangxiang Zeng}
\authornote{Corresponding Author}
\affiliation{%
  \institution{College of Computer Science and Electronic Engineering, Hunan University}
  \country{China}
  }
\email{xzeng@hnu.edu.cn}
\renewcommand{\shortauthors}{Trovato and Tobin, et al.}

\begin{abstract}
  Knowledge graph completion (KGC) aims to alleviate the inherent incompleteness of knowledge graphs (KGs), which is a critical task for various applications, such as recommendations on the web. Although knowledge graph embedding (KGE) models have demonstrated superior predictive performance on KGC tasks, these models infer missing links in a black-box manner that lacks transparency and accountability, preventing researchers from developing accountable models. Existing KGE-based explanation methods focus on exploring key paths or isolated edges as explanations, which is information-less to reason target prediction. Additionally, the missing ground truth leads to these explanation methods being ineffective in quantitatively evaluating explored explanations. To overcome these limitations, we propose KGExplainer, a model-agnostic method that identifies connected subgraph explanations and distills an evaluator to assess them quantitatively. KGExplainer employs a perturbation-based greedy search algorithm to find key connected subgraphs as explanations within the local structure of target predictions. 
  To evaluate the quality of the explored explanations, KGExplainer distills an evaluator from the target KGE model. By forwarding the explanations to the evaluator, our method can examine the fidelity of them. Extensive experiments on benchmark datasets demonstrate that KGExplainer yields promising improvement and achieves an optimal ratio of 83.3\% in human evaluation.

\end{abstract}



\keywords{Model Transparency, Model Explanation, Knowledge Graph Embedding, Knowledge Graph Completion}

\maketitle


\section{Introduction}
\begin{figure}
  \includegraphics[width=0.85\columnwidth]{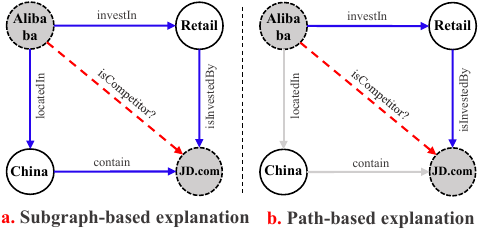}
  \caption{Here is an example of subgraph- and path-based \textcolor{blue}{explanations}. When considering the target prediction (\textcolor{red}{dashed line}) $\left<Alibaba,\mathbf{isCompetitor},JD.com\right>$, the path-based explanation $\left<Alibaba,\mathbf{investIn}, Retail, \mathbf{isInvestedBy}, JD.com\right>$ cannot always conclude
  the fact $\left<Alibaba,\mathbf{isCompetitor},JD.com\right>$ due to the location of them is confusing. In contrast, the subgraph-based explanation introduces an additional fact: $Alibaba$ and $JD.com$ are both located in $China$. The fact allows for an accurate deduction of the target prediction.
  }
  \label{fig:intro}
\end{figure}
Knowledge graph completion (KGC) aims to infer missing facts and tackles the incompleteness of knowledge graphs (KGs)~\cite{akrami2020realistic}, which has been widely used to support various applications including recommendation~\cite{wang2019knowledge,yang2022knowledge,wang2022graph}, sponsored search~\cite{lin2022investigating}, and drug discovery~\cite{kg_mtl,bang2023biomedical,pan2022deep}. 
Most KGC models utilize knowledge graph embedding (KGE) methods to map KG elements into multi-dimensional vectors~\cite{nguyen2023link,song2023xgcn}. These vectors are commonly used as features for downstream tasks and input into scoring functions for predicting missing links. KGE methods have been shown to achieve more powerful predictive performance compared to traditional models and can be scaled to large graphs~\cite{ali2021bringing,zheng2020dgl}.
However, current KGE methods make black-box predictions without providing explanations
, hindering their deployment in risk-sensitive scenarios~\cite{wang2022graph,zhang2023page}.

To address the aforementioned limitation, traditional rule-based methods leverage the known facts within the KG to extract possible logical rules, usually represented as paths, to explain the predicted fact
for knowledge graph completion~\cite{zhang2019iteratively,sadeghian2019drum,arakelyan2021complex}.
However, the rule mining-based KGC models are limited in their predictive performance.
To achieve comparable performance to embedding-based KGC models while simultaneously providing explanations, adversarial modifications are adopted to find the key facts, which identify the facts to include or exclude from the KG and monitor their prediction score change in the perturbed KG~\cite{pezeshkpour2019investigating,rossi2022explaining,betz2022adversarial}. However, these methods focus on providing a single fact or multiple isolated facts~\cite{zhao2023ke,baltatzis2023kgex} (i.e., discrete subgraph)
as explanations for a model prediction
, which do not represent coherent reasoning chains and are insufficient to
completely explain why the model makes such a prediction.
Reinforcement learning is adopted to address the previous issue by directly learning paths as prediction explanations, which balances the model's faithfulness and explainability~\cite{xian2019reinforcement,bhowmik2020explainable,jiang2023rcenr}. 
Although this method
can obtain valid explainable paths
, they are not expressive enough to provide explanations in the form of subgraphs, which are essential for predictions on complex KGs.
For example, as shown in Figure~\ref{fig:intro},
\textit{Alibaba} is identified as a competitor of \textit{JD.com}  over a complex KG\footnote{Alibaba and JD.com are online retail companies in China.}.
The path-based explanation $\left<Alibaba,\mathbf{investIn}, Retail, \mathbf{isInvestedBy}, JD.com\right>$ cannot always conclude the fact $\left<Alibaba,\mathbf{isCompetitor}, JD.com\right>$ due to the location of them is confusing. In contrast, the subgraph-based explanation of the target fact is informative with two additional facts: $Alibaba$ and $JD.com$ both lie in $China$. These facts can effectively conclude
that $Alibaba$ is $JD.com$'s competitor. Additionally, for the identified explanations, we cannot directly verify their rationality due to the ground truth is unavailable. Previous methods focused more on model faithfulness~\cite{pezeshkpour2019investigating,xian2019reinforcement} or manual verification of small-scale facts case by case~\cite{zhang2021explaining}, ignoring the quantitative assessment of model explainability.

Based on the above observations, we propose KGExplainer, a post-hoc explainability model performed on the trained KGEs, to search for connected subgraph-level explanations
and provide an effective strategy for evaluating the explanations quantitatively.
We first perform a model-agnostic greedy search approach to identify important subgraphs and distill a subgraph evaluator from target KGEs to measure the correlation between subgraphs and predictions. Then we evaluate the fidelity~\cite{wu2023explaining} of the identified subgraphs by measuring the performance correlation between the original KGC model that has access to the whole KG and the evaluator only visible to the explanation subgraphs.

In summary, the contributions of KGExplainer include:
\begin{itemize}[leftmargin=*]
    \item 
    We approach the explanations of KGE-based KGC models from a novel perspective by emphasizing and evaluating connected subgraph explanations.
    \item We present KGExplainer, a model-agnostic perturbation-based greedy search algorithm that can extract explanations with critical connected subgraph patterns for knowledge graph completion in a coherent way. 
    \item 
    We propose an evaluation strategy to quantitatively evaluate the effectiveness of subgraph-based explanations by distilling an evaluator to examine their fidelity.
    \item 
    Extensive experiments and human evaluation on widely used datasets demonstrate the effectiveness and efficiency of KGExplainer for exploring human-understandable explanations.
\end{itemize}

\section{Related Works}
\subsection{Knowledge Graph Completion}
Knowledge graph completion (KGC) aims to address the invariable incompleteness of knowledge graphs (KGs) by identifying missing interactions between entities. 
Previous rule-based methods~\cite{sadeghian2019drum,arakelyan2021complex} mine logical rules iteratively based on pre-trained embeddings of KGEs
during the training process.
These methods are limited to the predictive performance and scalability of KGC tasks on KGs. 
Knowledge graph embedding (KGE) models map KG elements to multi-dimensional vectors and define a scoring function to infer new facts, which have shown superior performance over rule-based methods
for predicting missing links~\cite{zhang2022rethinking,wang2023hyconve,gregucci2023link}. 
Specifically, the translational models, TransE~\cite{bordes2013translating} and its extensions~\cite{wang2014knowledge,lin2015learning}, represent relations and entities as embedding vectors and treat the relation as the translation from head entities to tail entities. In another aspect, the bilinear models, RESCAL~\cite{nickel2011three} and its variants~\cite{trouillon2016complex,yang2014embedding}, represent relations with matrices and combine the head and tail entities by sequentially multiplying head embedding, relation matrix, and tail embedding.
To effectively infer various relation patterns or properties
(e.g., symmetry and inversion) over complex KGs, RotatE~\cite{sun2019rotate}
represents
each relation as a rotation operation
from the source entity to the target entity in the complex vector space. Although these KGE models are successful in predicting unknown facts, they are limited in lacking transparency and accountability, which block researchers from developing trustworthy models. To address the above limitations, we propose KGExplainer to explore subgraph-based explanations for KGE-based KGC models.

\subsection{Explainability in Knowledge Graph Completion}


To increase the transparency of models in KGC tasks, researchers have developed various explanation models~\cite{wang2022graph,huang2022trustworthy,zhang2023page,yao2023path}. Most explanation models for 
KGC can be categorized into rule-based, fact-based, and path-based methods. 
Learning rules as the explanations from KGs has been studied extensively in inductive logic programming~\cite{galarraga2013amie,qu2020rnnlogic}. RuLES~\cite{ho2018rule} extended rule learning by exploiting probabilistic representations of missing facts computed by a pre-trained KGE model.
DRUM~\cite{sadeghian2019drum} proposed a scalable and differentiable approach to mine first-order logical rules from KGs for unseen entities. The mining rules are human-understandable and can be used to deduce predicted facts and provide logical explanations~\cite{zhang2019iteratively}. While rule-based models are native explainability, they often fall short in achieving excellent predictive performance and scaling to complex KGs. 
To improve the explainability of high-accuracy KGE methods,
CRIAGE~\cite{pezeshkpour2019investigating}, Kelpie~\cite{rossi2022explaining}, and KE-X~\cite{zhao2023ke} search for the most important edges (i.e., isolated facts) to explain the target prediction by approximately evaluating
the impact of removing an existing fact from the KG on prediction score. However, these methods only provide a single fact or multiple discrete facts as explanations, thus lacking enough information to completely explain a model prediction. To explore more effective explanations, PGPR~\cite{xian2019reinforcement} and ELEP~\cite{bhowmik2020explainable} have adopted a reinforcement learning approach to predict tail entities and take the reasoning path between the head and tail entities as the explanation. PaGE-Link~\cite{zhang2023page} proposed a path-based graph neural network explanation for the KGC tasks, providing human-understandable explanations in a learnable way. As shown in Figure~\ref{fig:intro}, although these methods can reason explanatory paths, they are invalid in explaining predictions over complex KGs. Additionally, the explainable ground truth is unavailable, leading to it being difficult for previous methods to evaluate the explanations quantitatively. To address these limitations, we propose KGExplainer to find subgraph-based explanations that most influence the target prediction and distill an evaluator to assess the explored explanations quantitatively.

\section{Preliminary}
\subsection{KGE-based Knowledge Graph Completion}
We define a knowledge graph (KG) as a labeled directed graph $\mathcal{G}=(\mathcal{V},\mathcal{R},\mathcal{E})=\{(h, t, t)| h,r \in \mathcal{V}, r\in \mathcal{R}, (h,t)\in \mathcal{E}\}$
where each triple represents a relation $r$ between the head entity $h$ and tail entity $t$. Most KGs are incomplete and many unknown interactions are undiscovered. Knowledge graph completion (KGC) leverages existing facts in the KG to infer missing ones, which alleviates the incompleteness of the KG.
Currently, the most popular KGC method adopts deep learning techniques to learn the embeddings of KG elements (i.e., vectorized representations of entities and relations). Intuitively, knowledge graph embeddings (KGEs) contain
the semantic information of the original KG and can be used to predict new links. Generally, the KGE-based models define a score function $\phi$ to estimate the plausibility of triples 
and optimize the vectorized embeddings by maximizing the scores of existing facts. For the incomplete triple $\left<h,r,?\right>$, the KGE models find the missing tail entity $t$ with best scores as follows:
\begin{equation}
    t=\operatorname*{argmax}_{v\in\mathcal{V}}\phi(h,r,v).
\end{equation}
The head entity prediction for the unknown links like $\left<?,r,t\right>$ can be defined analogously. In the following sections, we mostly refer to tail entity prediction for simplicity. All our methods can be applied to
head entity prediction.

\subsection{Enclosing Subgraph of Target Prediction}
Based on GraIL~\cite{teru2020inductive}, when given a KG $\mathcal{G}$ and a link $\left<h,r,t\right>$ for tail prediction, we extract an enclosing subgraph surrounding the target link to represent the relevant patterns for model's decision. Initially, we obtain the $k$-hop neighboring entities $\mathcal{N}_{k}(h)=\{s|d(h,s)\leq k\}$ and $\mathcal{N}_{k}(t)=\{s|d(t,s)\leq k\}$ for both $h$ and $t$, where $d(\cdot,\cdot)$ represents
the shortest path distance between the entity pair on $\mathcal{G}$. We then obtain the set of entities $V=\{s|s\in \mathcal{N}_{k}(h)\cap \mathcal{N}_{k}(t)\}$ as vertices of the enclosing subgraph. Finally, we extract the edges $E$ linked by the set of entities $V$ from $\mathcal{G}$ as the $k$-hop enclosing subgraph  $g=(V, E)$.

\section{The KGExplainer Framework}
\begin{figure}
  \includegraphics[width=0.98\columnwidth]{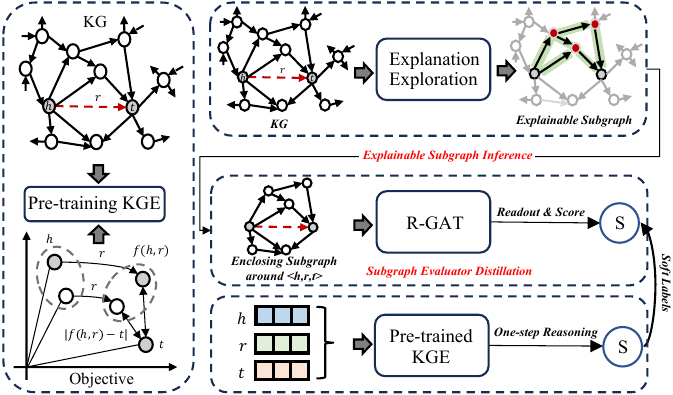}
  \caption{The KGExplainer framework comprises three modules: (1) Pre-training KGE models on the target KG
  ; (2) Exploring subgraph-based explanations for pre-trained KGE models by greedy search
  ; (3) Distilling a subgraph structure evaluator from pre-trained KGE to assess the explored explanations quantitatively. }
  \label{fig:method}
\end{figure}
\subsection{Problem Formulation}
Given a fact $\left<h,r,t\right>$,
KGExplainer investigates
the explanation that intuitively consists of the most critical subgraph patterns featuring $h$ that allow KGE models to predict tail $t$. For instance, when explaining why the top ranking tail for the missing fact $\left<Donald\_Trump, nationality,?\right>$ is $USA$, 
KGExplainer searches for the smallest key subgraph between \textit{Donald\_Trump} and \textit{USA}, allowing the model to predict the same ranking of \textit{USA} based on the key subgraph. Specifically, KGExplainer aims to find a subgraph $g_{key}$ that is most influential to the target prediction $\left<h,r,t\right>$, and removing the other irrelevant facts have no effect on the model's prediction. We can define this process as follows:
\begin{equation}
    g_{key}=\operatorname*{argmin}_{\Tilde{g}\in g_{sub}}\Delta(f(g), f^\prime(\Tilde{g})),
\end{equation}
where $g$ is the original enclosing subgraph $g$
, $g_{sub}$ denotes the all possible connected subgraphs from $h$ to $t$ within $g$, and $f(\cdot)$ and $f^\prime(\cdot)$ represent the original and retrained KGC models.

\subsection{Overview of KGExplainer}
KGExplainer is a model-agnostic and post-hoc framework that can be designed for explaining and evaluating
KGE-based KGC models.
Given a prediction from the pre-trained KGE models, KGExplainer identifies the smallest subgraph to preserve the prediction performance.
As illustrated in Figure~\ref{fig:method}, KGExplainer comprises three main components: \textit{KGE Pre-training}, \textit{Explanation Exploration} and \textit{Subgraph Evaluator Distillation}. Specifically, KGExplainer first pre-trained a KGE (e.g., RotatE) as the embedding-based KGC model for the subsequent exploration and evaluation as presented in Section~\ref{sec:pre-train}.
When given a pre-trained KGE model, KGExplainer uses a greedy search algorithm based on the enclosing subgraph of the target fact to find the key subgraph as an explanation as shown in Section~\ref{exploration}.
Meanwhile, KGExplainer distills a subgraph neural network to model the semantic and structural topology of enclosing subgraphs surrounding the predicted links
by target KGE as explained in Section~\ref{distill}. The distilled subgraph representation model provides insights into evaluating the relevance between a target fact and a subgraph. Therefore, we use the subgraph neural network as an evaluator to assess whether the identified key patterns can preserve the target prediction by forwarding it into the distilled model.

\subsection{Pre-training KGE} \label{sec:pre-train}
In this paper, we developed the KGExplainer as a post-hoc explanation method for KGE-based KGC models by exploring complex subgraph patterns. 
To effectively learn semantic knowledge within the complex KG, we adopt the state-of-the-art RotatE~\cite{sun2019rotate}
as the target KGE. Specifically, given a triple $\left<h,r,t\right>$, we adopt a score function and expect that the embedding of head entity $h$ is similar to the embedding of tail entity $t$ by rotating it over the relation $r$. 
By maximizing the scores of positive triples and minimizing the scores of negative ones, we can obtain the pre-trained entity and relation embeddings. 
Theoretically, KGExplainer is model-agnostic and can be applied to any KGE-based model with a score function. 
We designed experiments to verify the effectiveness of KGExplainer on various KGE models, the details can be found in Section~\ref{param:kges}.
\begin{figure}
  \includegraphics[width=0.95\columnwidth]{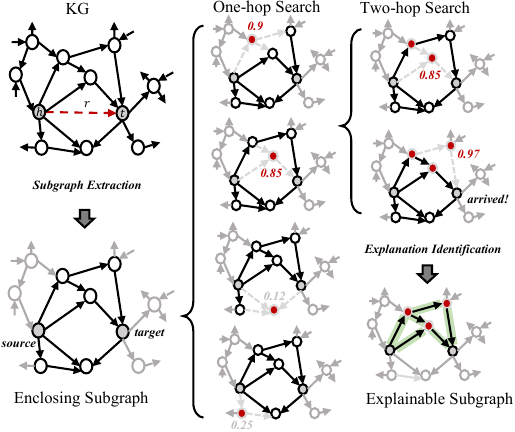}
  \caption{The details of exploring explanations. KGExplainer searches top $n=2$ key entities per hop from the source to target entities greedy within the enclosing subgraph, and then extracts the key subgraph by the identified entities.}
  \label{fig:exploration}
\end{figure}
\subsection{Explanation Exploration}\label{exploration}
\begin{algorithm}[t]
\SetAlgoLined
\SetAlgoNoEnd
\SetKwInOut{Data}{Input}
\SetKwInOut{Result}{Output}
\Data{Target prediction $\left<h,r,t\right>$; Enclosing subgraph $g$; Prediction score $s$;
Subgraph extracting function $F_{sub}$;
Score function $\phi$;
Number of entities retrieved per hop $n$;
}
\Result{A key subgraph $g_{key}$ that is most critical to the target prediction $\left<h,r,t\right>$.}
Initialize a set of visited entities $S_{visit}$ as $\{h\}$\;
Initialize a set of key entities $S_{key}$ as $\{h\}$\;
Initialize a queue of searching entities $S_{search}$ as $\{h\}$\;
\While{$S_{search}$ is not empty}
{
  Initialize $H$ as an empty max-Heap\;
  $e_{searching} \gets S_{search}.pop()$\;
  \If {$t\in \mathcal{N}(e_{searching})$} {
  $S_{key} \gets S_{key}\cup \{t\}$ \;
  break\;
  }
  \For{$v \in \mathcal{N}(e_{searching})$} {
  \If{$u \notin S_{visit}$} {
  $S_{visit} \gets S_{visit}\cup \{u\}$ \;
    Create $g^\prime$ by removing $u$ from $g$ and finetune the KG embeddings of $g^\prime$\;
    Get new embeddings of $h$, $r$, $t$ as $\mathbf{h^\prime}$, $\mathbf{r^\prime}$, $\mathbf{t^\prime}$\;
      $s^\prime \gets \phi(\mathbf{h^\prime},\mathbf{r^\prime},\mathbf{t^\prime})$\;
      $\delta_v \gets s - s^\prime$\;
      Put $v$ into $H$ with its corresponding value $\delta_v$\;
  } 
  }
  \For{$i \gets 1$ \textbf{to} $n$} {
    Pop $v$ with the maximum $\delta_v$ from $H$ and put it into $S_{key}$ and $S_{search}$\;
  }
}
$g_{key}\gets F_{sub}(g, S_{key})$\;
\textbf{return} $g_{key}$\;
\caption{Greedy search for subgraph explanation}
\label{algo:greedy-entity}
\end{algorithm}
\subsubsection{Greedy Search for Key Subgraph.}
As shown in Figure~\ref{fig:exploration}, given a predicted link $\left< h, r, t\right>$ for the prediction of tail entity $t$, we developed a greedy search algorithm based on a perturbation mechanism to search for crucial subgraph patterns as the explanation. We first extract the enclosing subgraph $g$ around the target link, which contains informative interactions for such prediction. To identify the most important subgraph influencing the prediction within $g$, we search all entities in $g$ and measure their importance to capture key ones. Specifically, we remove each entity and its adjacent from $g$ and retrain\footnote{Retraining KGE over the entire KG is way too costly and we discuss an efficient strategy to tackle this issue in Section~\ref{re-training}} the target KGE on the perturbed subgraph $g^\prime$.
Then we measure the score change of the target link using
the original and retrained embedding as the entity importance. The more the score changes, the more critical an entity is. To ensure that the key entities obtained from the $k$-hop enclosing subgraph $g$
lead to a single connected component, we search the key entities from the head $h$ and tail $t$ hop by hop.
We first remove each 1-hop neighbor entity of $h$ and measure their importance to identify the top important entities.
After getting the top-$n$ important entities $\mathrm{v}_{1}^n$ from 1-hop neighbor entities, we 
then search the one-hop neighboring entities of $\mathrm{v}_{1}^n$ analogously for important $2$-hop neighbors $\mathrm{v}_{2}^n$. 
We repeat this one-hop search process until reaching the tail entity $t$ of the target link. 
As a result, we capture a set of the $k$-hop key neighboring
entities $\mathrm{v}^n = \{\mathrm{v}_{1}^n,\cdots,\mathrm{v}_{k}^n\}$ and construct a connected
subgraph as the explanation for the target prediction.
Algorithm~\ref{algo:greedy-entity} formally describes the process.

\subsubsection{Retraining Strategy for Importance Assessment.}\label{re-training}
To effectively obtain the score changes of target facts based on the KGE models, we designed a retraining strategy to capture the new embedding on the perturbed KG. Since each entity perturbation only makes minor changes over the entire KG, 
retraining the embeddings over the revised KG is infeasible when $\mathcal{G}$ has millions of entities and facts. To focus on the relevant patterns and reduce the overhead of the retraining, we introduce a subgraph-based retraining strategy. We extract the enclosing subgraph $g$ of the target fact $\left< h, r, t\right>$ and expand $g$ by including the 1-hop neighbors $\mathcal{N}_{extend}$ of the existing entities, which constructs a new subgraph $g_{extend}$. Then we retrain the embeddings of the KG elements on $g_{extend}$ to make the calculation of score changes effective. To keep the overall entity and relation embeddings of $g$ within the same representation space of $\mathcal{G}$, we alleviate the following constraints: (1) We initialize the embeddings of the entities in $g_{extend}$ with the pre-trained embeddings from $\mathcal{G}$; (2) We fix the relation embeddings in $g_{extend}$ as the same as those in $\mathcal{G}$; (3) We fix the embeddings of entities $v\in \mathcal{N}_{extend}$. By using the retraining strategy, KGExplainer can effectively capture the target fact's score change and efficiently search for key subgraphs. More details can be found in Appendix~\ref{retrain}.

\subsection{Subgraph Evaluator Distillation}\label{distill}
\subsubsection{Why Subgraph Evaluator?}
Evaluating the quality of identified explanation subgraphs is essential. However, the ground truth explanations are unavailable, making it difficult to achieve that objective. Following~\cite{chen2018learning}, the explanatory subgraphs can be viewed as the discriminative information coherent in the model knowledge.
Thus we propose a post-hoc evaluation strategy to assess the performance of the original KGC model only visible to the explanation subgraphs. That is, we feed the explanation subgraphs to the model and examine how well they recover the target predictive rankings. 
However, pre-trained KGEs implicitly have access to the entire graph and we cannot restrict the prediction of a KGE-based KGC model to a subgraph.
To address this limitation, we design a subgraph evaluator by distilling a
relational graph neural network (GNN) from pre-trained KGEs for KGC. During training, the GNN is optimized to predict the facts from their complete enclosing subgraphs. Once trained, it can be used to evaluate the degree of performance preservation with subgraph explanations. 
Given a pre-trained KGE-based KGC model $\phi(h, r, t)$,
we aim at training a subgraph evaluator to replicate the predictive performance of the target KGE and evaluate the subgraph structure of explanation.

\subsubsection{How to get Subgraph Evaluator?}
Given a link $\left<h,r,t\right>$, we develop a score function $Z(h,r,t)$ with a $L$-layer relational graph attention neural network (R-GAT) on its enclosing subgraph $g$ to score it based on the local substructure representation. The update function of the entities in the
$l$-th layer is defined as:
\begin{equation}
\begin{aligned}\mathbf{x}_i^l&=\sum_{r\in\mathcal{R}}\sum_{j\in\mathcal{N}_r(i)}\alpha_{(i,r,j)}\mathbf{W}_r^l\Phi(\mathbf{e}_r^{l-1},\mathbf{x}_j^{l-1}),
\\\alpha_{(i,r,j)}&=\mathrm{sigmoid}\left(\mathbf{W}_1\left[\mathbf{x}_i^{l-1}\oplus\mathbf{x}_j^{l-1}\oplus\mathbf{e}_r^{l-1}\right]\right),\end{aligned}
\end{equation}
where $\mathbf{e}_r^{l}$ and $\mathbf{x}_i^{l}$ represent the $l$-th layer embeddings of relation $r$ and entity $i$. $\alpha_{(i,r,j)}$
and $\mathcal{N}_r(i)$ denote
the weight and neighbors of entity $i$ under the relation
$r$, respectively. $\oplus$ is the concatenation operation. $\mathbf{W}_r^l$ represents the transformation matrix of relation $r$, and $\Phi$ is the aggregation operation to fuse the hidden features of entities and relations. Finally, we obtain the global representation $\mathbf{X}_{sub}$ of the subgraph $g$ as follows:
\begin{equation}
    \mathbf{X}_{sub}=\frac1{|V|}\sum_{i\in V}f(\mathbf{x}_i^L),
\end{equation}
where $V$ is the entity set of the enclosing subgraph $g$. After obtaining the global representation $\mathbf{X}_{sub}$, we define a linear layer to score the target link as follows:
\begin{equation}
    Z(h,r,t) = \mathbf{W}^{T}\mathbf{X}_{sub},
\end{equation}
where $\mathbf{W}^{T}$ is a transform matrix.

To optimize the subgraph evaluator,
inspired by~\cite{gou2021knowledge},
we adopt the insights of knowledge distillation and consider the predictions of the pre-trained KGE-based KGC model as soft labels. Meanwhile, to enhance the model's ability to rank candidate entities, we introduce a regularization term that encourages maximizing the margin between the scores for positive and negative samples.
The objective for the optimization of this evaluator is defined as follows:
\begin{equation}
\min\sum_{(h,r,t)\in\mathcal{G}}\|\phi(h,r,t)-Z(h,r,t)\|^2+\lambda L(h,r,t),
\end{equation}
where the first term is a score
alignment objective.
The second term is a pair-wise loss
, which is defined by negative sampling:
\begin{equation}
    L(h,r,t)=-Z(h,r,t)+\frac{1}{N}\sum_{n=1}^N Z(h,r,v_n),
\end{equation}
where $N$ denotes the number of negative samples and $v_n$ is uniformly drawn from $V$. 
By minimizing the above optimization objective, we can obtain an effective subgraph scoring function with powerful predictive performance and evaluation capabilities for subgraph-based explanations.

\subsection{Computation Complexity of KGExplainer}
The computation complexity of finding explanations using KGExplainer is related to two parts: 1) the size of enclosing subgraph $g$ and 2) the number of entities that require importance evaluation. 

\newtheorem{prop}{Proposition}
\begin{prop}
Given $\mathcal{G}=(\mathcal{V},\mathcal{R},\mathcal{E})$, a predicted fact $\left<h,r,t\right>$, the maximum length of paths $L$, and the number of paths $N$, the size of $g$ is $O(\frac{LN|\mathcal{V}|}{|\mathcal{E}|})$ and the size of $\mathcal{N}_{extend}$ is $O(LN)$.
\end{prop}
Given $h$ and $t$, the maximum number of edges within paths is $L*N$. The probability of an entity $v \in \mathcal{V}$ in an edge $e \in \mathcal{E}$ is $\frac{2|\mathcal{V}|}{|\mathcal{E}|}$. Thus, the size of $g$ is bounded by $LN*\frac{2|\mathcal{V}|}{|\mathcal{E}|}$, i.e., $O(\frac{LN|\mathcal{V}|}{|\mathcal{E}|})$. The number of entities in $\mathcal{N}_{extend}$ is $\frac{2LN|\mathcal{V}|}{|\mathcal{E}|} * \frac{2|\mathcal{E}|}{|\mathcal{V}|}$, i.e., $O(LN)$.
\begin{prop}
Given $\mathcal{G}=(\mathcal{V},\mathcal{R},\mathcal{E})$, a predicted fact $\left<h,r,t\right>$, the maximum length of paths $L$ and the maximum number of entities per hop in an explanation $K$.
The maximum number of entities to evaluate is $(1+(L-2)K)(\frac{2|\mathcal{E}|}{|\mathcal{V}|})$.
\end{prop}
Following Algorithm~\ref{algo:greedy-entity}, the average number of entities to evaluate is $d$ in the first hop, $0$ in the last hop, and $K * d$ in the intermediate hops, where $d=\frac{2|\mathcal{E}|}{|\mathcal{V}|}$ is the average degree of an entity. Thus, the total number of entities to evaluate is $(1+(L-2)K)(\frac{2|\mathcal{E}|}{|\mathcal{V}|})$.

Here, we conclude that the average computation complexity of finding the explanation for a predicted fact $\left<h,r,t\right>$ is only related to the maximum length of paths $L$, the maximum number of paths $N$, and the average degree of an entity in KG, which is not related to the size of $\mathcal{G}$. Thus, in theory, KGExplainer can easily scale to large KGs.
\section{Experiment}\label{sec:experiment}
In this section, we evaluate the proposed KGExplainer\footnote{Sources are available at xxx}
by considering the following \textit{key} research questions.
\begin{itemize}[leftmargin=*]
    \item \textbf{Q1} Does KGExplainer have similar predictive performance with the target KGE models?
    \item \textbf{Q2} Is KGExplainer more effective than other fact- and path-based methods for exploring explanations?
    \item \textbf{Q3} Can KGExplainer work well with different sizable explainable subgraphs and various KGE methods?
    \item \textbf{Q4} Does KGExplainer produce sensible explanations?
    \item \textbf{Q5} Is KGExplainer efficient in exploring explanations?
\end{itemize}
\subsection{Datasets}
Similar to \cite{sadeghian2019drum}, we conduct experiments on three widely used datasets\footnote{\url{https://github.com/alisadeghian/DRUM/tree/master/datasets}} for the KGC task. Specifically, \textbf{WN-18}~\cite{transe} is a database with complex patterns featuring lexical relations between words extracted from WordNet~\cite{wornet}. The \textbf{Family-rr}~\cite{sadeghian2019drum} is a dataset that contains the bloodline relationships between individuals of various families. \textbf{FB15k-237}~\cite{transe} is a subset of Freebase~\cite{freebase}, which is
a large-scale knowledge graph containing general facts. For every dataset, we split it into training and test subsets for the following evaluations of faithfulness and explainability. The statistics of all datasets are shown in Table~\ref{tab:statistic}. \#train and \#test are the number of triples on the training and testing sets.

\subsection{Evaluation Settings}
\begin{table}[t]
\caption{The statistics of datasets.}
\begin{tabular}{lcccc}
\toprule
\textbf{Dataset}   & \textbf{\#entity} & \textbf{\#relation} & \textbf{\#train} & \textbf{\#test} \\ \midrule
\textbf{WN-18}     & 40,943            & 18                  & 35,354                   & 2,250                   \\
\textbf{Family-rr} & 3,007             & 12                  & 5,868                    & 2,835   \\ 
\textbf{FB15k-237} & 14,541            & 237                 & 68,028                   & 9,209                   \\ \bottomrule               
\end{tabular}

\label{tab:statistic}
\end{table}
\subsubsection{Faithful Metrics.}
The evaluation
of KGC models is performed by running head and tail prediction on each triple in the test set. For each prediction, the ground truth is ranked against all the other entities $\mathcal{V}$. Without loss of generality, for tail prediction, this corresponds to:
\begin{equation}
    tailRank(h,r,t)=|\{v\in\mathcal{V}|\phi(h,r,v)>=\phi(h,r,t)\}|.
\end{equation}
Following~\cite{kadlec2017knowledge}, we adopt
$Hits@N$ to quantify the predictive performance of KGE and distilled models.
$Hits@N$ is the fraction of ranks $S_{rank}$ with value $\leq N$:
\begin{equation}
    Hits@N=\frac{|\{s_i\in S_{r}:s_i\leq N\}|}{|S_{r}|},
\end{equation}
where $s_i$ is the ranking of $i$-th candidate entity and $S_{r}$ is a set of rankings for all candidate entities.
Following the method proposed by~\citeauthor{kadlec2017knowledge}, we focus on the $Hits@1$ metric to better highlight model differences.

\subsubsection{Explanation Metrics.}
As the ground truth explanations of KGC tasks are usually missing, it is not easy to quantitatively evaluate the quality of explanations.
We consider explanations to be sufficient if they allow preserving the KGC performance~\cite{akrami2020realistic,wang2021towards,wu2023explaining}. Thus we define the $Recall@N$ and $F1@N$ as the metrics to evaluate the effectiveness of explored explanations. Specifically, $Recall@N$ measures the fidelity of explanatory subgraphs by feeding
them into the distilled model and examining how well they
recover the target predictive rankings. Given the rankings of tested facts by forwarding their explanations to the distilled evaluator, we calculate the $Recall@N$ as follows:
\begin{equation}
    Recall@N=\frac{|\{s_i\in S_{r}:s_i\leq N\}\cap \{s_j\in S_{r}^{exp}:s_j\leq N\}|}{|\{s_i\in S_{r}:s_i\leq N\}|},
\end{equation}
where $S_{r}^{exp}$ is the ranking set of explanations and $S_{r}$ denotes the original predictive rankings of them. We can assess the trustworthiness 
of explanations by calculating the recovery rate of the rankings $S_r$.
Meanwhile, the subgraph-based explanations are critical in the precision of rankings (i.e., $Hits@N$), to comprehensively evaluate the explainability of KGExplainer, we define $F1@N$ to balance the recall and precision of the explored explanations:
\begin{equation}
\begin{aligned}
    &Hits@N = \frac{|\{s_j\in S_{r}^{exp}:s_j\leq N\}|}{|S_{r}^{exp}|},\\
    &F1@N=\frac{2\cdot Recall@N\cdot Hits@N}{Recall@N + Hits@N}.
\end{aligned}
\end{equation}
Similar to the evaluation of faithfulness, we consider $Recall@1$ and $F1@1$ for the qualitative metrics to emphasize the performance discrepancy.



\begin{table}[]
\caption{The predictive performance of KGE and the distilled models on $Hits@1 (\%)$ metric over the three datasets.}
\label{tab:faith}
\begin{tabular}{llll}
\toprule
\multicolumn{1}{c}{\textbf{Methods}} & \multicolumn{1}{c}{\textbf{WN-18}} & \multicolumn{1}{c}{\textbf{Family-rr}} & \multicolumn{1}{c}{\textbf{FB15k-237}} \\ \midrule
TransE                      & 29.50 {\footnotesize(0.32)}                    & 12.63 {\footnotesize(0.09)}                                & 41.86 {\footnotesize(0.43)}                                \\
KGExp-TransE                      & 27.40 {\footnotesize(0.23)}                   & 15.77 {\footnotesize(0.11)}                                 & 40.99 {\footnotesize(0.31)}                                 \\ \midrule
DistMult                    & 9.450 {\footnotesize(0.15)}                             & 8.680 {\footnotesize(0.27)}                         & 10.26 {\footnotesize(0.21)}                                \\
KGExp-DistMult                      & 16.65 {\footnotesize(0.16)}                    &  9.881 {\footnotesize(0.14)}                               & 13.09 {\footnotesize(0.32)}                                 \\ \midrule
RotatE                      & 29.38 {\footnotesize(0.19)}                            & 39.08 {\footnotesize(0.45)}                       & 42.10 {\footnotesize(0.51)}  \\
KGExp-RotatE                      & 29.47 {\footnotesize(0.22)}                    & 35.79 {\footnotesize(0.06)}                                 & 39.28 {\footnotesize(0.29)}                                 \\ \bottomrule

\end{tabular}
\end{table}

\subsection{Baselines}
To verify the performance of KGExplainer in exploring explanations, we compare it against various baselines as follows:
\begin{itemize}[leftmargin=*]
    \item \textbf{Fact-based Methods:} \textbf{CRIAGE}~\cite{pezeshkpour2019investigating} identified the most important fact to explain a prediction by approximating the impact of removing an existing fact from the KG.
    \textbf{Kelpie}~\cite{rossi2022explaining} identified multiple isolated facts as the reasoning evidence and can be applied to any embedding-based KGC models.
    \item \textbf{Rule- or Path-based Methods:} \textbf{DRUM}~\cite{sadeghian2019drum} is a scalable and differentiable model for mining logical rules as path-based explanations from KGs while
    predicting unknown links.
    \textbf{ELEP}~\cite{bhowmik2020explainable} utilized a reinforcement learning approach to find reasoning paths between the head and tail entities of the target prediction.
    \textbf{PaGE-Link}~\cite{zhang2023page} can generate explanations as paths connecting the entity pair based on subgraphs in a learnable way.  
\end{itemize}
Furthermore, we design variants of KGExplainer to verify its effectiveness on different embedding-based KGC models. For convenience,
we denote \textbf{KGExplainer} as \textbf{KGExp} in tables.
\begin{itemize}[leftmargin=*]
    \item \textbf{KGExp-Rand} randomly selected some connective paths between head and tail entities to construct an explainable subgraph, which allows an ablation study.
    \item \textbf{KGExp-TransE}, \textbf{KGExp-DistMult}, and \textbf{KGExp-RotatE} are variants of KGExplainer that
    consider TransE~\cite{transe}, DistMult~\cite{yang2014embedding}, and RotatE\cite{sun2019rotate} as the base KGE, respectively.
\end{itemize}
More implementation details of baselines refer to Appendix~\ref{appendix:setting}.
\subsection{Performance and Comparison}

In this section, we analyze the performance of KGExplainer from two aspects, faithfulness and explainability. 
The average
performance and standard deviation over five runs are reported in Table~\ref{tab:F1} and Table~\ref{tab:Recall1}. We show more details in Appendix~\ref{appendix:full_results}.

\subsubsection{Faithfulness Evaluation (Q1).}
An important goal of explainability models is to accurately and comprehensively represent the local decision-making structure of the target model. Specifically, in the context of KGE models, KGExplainer should strive to replicate their behavior as closely as possible~\cite{zhang2021explaining}. Thus, to answer \textbf{Q1}, we compare the predictive performance of KGExplainer and the target KGE models. As shown in Table~\ref{tab:faith}, we observe that KGExplainer achieves nearly the same performance on the KGC task compared with the target KGE models (e.g., the performance of KGExp-RotatE and RotatE on WN-18 dataset). Furthermore, we observe that RotatE overall achieves the best performance for KGC and similarly KGExp-RotatE shows superior predictive performance among the distilled models, which brings greater credibility than others, supporting that KGExplainer is faithful to target KGE and is reliable in replicating the target KGE. Thus, we chose the RotatE as the target KGE model and distilled a powerful evaluator based on the subgraph substructure around the target link to assess the explainability.

\subsubsection{Explainability Evaluation (Q2).}
\begin{table}[]

\caption{The $F1@1 (\%)$ performance of explanations over various datasets. The \textbf{boldface} denotes the highest score and the underline indicates the best baseline.}
\label{tab:F1}
\begin{tabular}{lccc}
\toprule
\textbf{Methods} & \textbf{WN-18} & \textbf{Family-rr} & \textbf{FB15k-237} \\ \midrule

CRIAGE           & 63.78 {\footnotesize (0.13)}             & 20.75 {\footnotesize (0.32)}         & 40.99 {\footnotesize (0.21)}                   \\
Kelpie           & 66.36 {\footnotesize (0.19)}             & 20.33 {\footnotesize (0.23)}        &  36.64 {\footnotesize (0.13)}                  \\ \midrule
DRUM             & 71.41 {\footnotesize (0.41)}            & 21.26  {\footnotesize (0.09)}       & 65.53 {\footnotesize (0.19)}                   \\
ELEP             & \underline{75.87} {\footnotesize (0.22)}                  & \underline{22.75} {\footnotesize (0.17)}              &  \underline{70.11} {\footnotesize (0.21)}                  \\
PaGE-Link        & 67.70 {\footnotesize (0.31)}                  & 21.89 {\footnotesize (0.13)}              &  66.12 {\footnotesize (0.14)}                    \\ \midrule
KGExp-Rand       & 31.22 {\footnotesize (0.53)}            & 19.16 {\footnotesize (0.26)}        &  55.17 {\footnotesize (0.23)}                  \\ 
KGExp-TransE     & 97.10 {\footnotesize (0.17)}            & 30.47 {\footnotesize (0.25)}       &   82.04 {\footnotesize (0.33)}                 \\
KGExp-DistMult   & 96.87 {\footnotesize (0.08)}            & \textbf{30.49} {\footnotesize (0.24)}        &  84.29 {\footnotesize (0.17)}                  \\ 
KGExp-RotatE     & \textbf{97.10} {\footnotesize (0.11)}            & 29.52 {\footnotesize (0.12)}        &  \textbf{83.65} {\footnotesize (0.14)}                  \\ \bottomrule     
\end{tabular}
\end{table}

It is essential to assess objectively how good an explanation is for KGExplainer. Unfortunately, similar to other explanation methods, evaluating explanations is made difficult by the impossibility of collecting \textit{ground truth} explanations. To fairly evaluate explanations without ground truth and answer \textbf{Q2}, we distill an evaluator and design two metrics $F1@1$ and $Recall@1$ to quantify how trusty an explanation is. As shown in Table~\ref{tab:F1} and Table~\ref{tab:Recall1}, we reported the $F1@1$ and $Recall@1$ results compared with baselines on three wide-used datasets, and more results can be found in Appendix~\ref{appdix:experiment}. Specifically, KGExplainer improves the $F1@1$ and $Recall@1$ by at least $21.23\%$ and $7.67\%$ respectively on the WN-18 dataset
and achieves $7.74\%$ and $4.77\%$ absolute increase over the best baseline on the Family-rr dataset. Meanwhile, the performance of KGExplainer on the FB15k-237 dataset has yielded $13.54\%$ and $4.19\%$ gains in $F1@1$ and $Recall@1$ compared with the baselines. 

Furthermore, we have the following observations: (1) Compared with the fact-based methods CRIAGE and Kelpie, the path-based explainability
methods DRUM, ELEP, and PaGE-link that utilize reasonable paths between source and target entities achieve better performance, which indicates that the connecting paths are important for the model explanation.
(2) Compared with the path-based explainability
models ELEP and PaGE-Link, subgraph-based KGExplainer (e.g., KGExp-RotatE) 
performs better than them on all datasets, demonstrating
that the subgraph-based explanations are more effective than paths and multiple disconnected facts. 
(3) For an ablation study,
KGExp-Rand, which adopts randomly sampled subgraphs connected to the source and target entities as the explanations, yields significantly inferior performance compared with other KGExplainer variants. This demonstrates that KGExplainer indeed identifies semantically meaningful subgraph explanations for model predictions, which are superior to randomly sampled subgraphs by a large margin. This may be because the KGExplainer based on target KGEs can explore reasonable subgraph patterns that are crucial to predicting the corresponding facts, and the randomly sampled subgraph may contain irrelevant information with missing key patterns to the predicted facts. The overall performance compared with the baselines demonstrates that KGExplainer is effective in exploring informative explanations and evaluating them.
\begin{table}[]

\caption{The $Recall@1 (\%)$ performance of explanations over various datasets. We mark the best score with bold font and mark the best baseline with underline.}
\label{tab:Recall1}
\begin{tabular}{lccc}
\toprule
\textbf{Methods} & \textbf{WN-18} & \textbf{Family-rr} & \textbf{FB15k-237} \\ \midrule
CRIAGE           & 86.69 {\footnotesize (0.10)}        & 28.57 {\footnotesize (0.33)}             & 68.23 {\footnotesize (0.43)}                   \\
Kelpie           & 88.31 {\footnotesize (0.28)}        & 27.62 {\footnotesize (0.39)}            &  60.01 {\footnotesize (0.27)}                  \\ \midrule
DRUM             & \underline{91.12} {\footnotesize (0.15)}        & 16.20 {\footnotesize (0.21)}            &  \underline{87.88} {\footnotesize (0.22)}                  \\
ELEP             & 89.97 {\footnotesize (0.11)}               & \underline{29.84} {\footnotesize (0.18)}                   & 85.58 {\footnotesize (0.29)}                   \\
PaGE-Link        & 85.36 {\footnotesize (0.25)}               & 27.14 {\footnotesize (0.31)}                   & 85.13 {\footnotesize (0.09)}                    \\ \midrule
KGExp-Rand       & 86.09 {\footnotesize (0.68)}        & 26.98 {\footnotesize (0.49)}            &  50.37 {\footnotesize (0.19)}                  \\
KGExp-TransE     & \textbf{98.79} {\footnotesize (0.22)}        & \textbf{34.61} {\footnotesize (0.43)}            &  \textbf{92.07} {\footnotesize (0.33)}                  \\
KGExp-DistMult   & 98.14 {\footnotesize (0.19)}        & 34.60 {\footnotesize (0.13)}            &  91.67 {\footnotesize (0.11)}                  \\
KGExp-RotatE     & 98.75 {\footnotesize (0.15)}        & 33.02 {\footnotesize (0.11)}            &  91.45 {\footnotesize (0.09)}                  \\ \bottomrule     
\end{tabular}
\end{table}
\begin{figure}[t]
  \includegraphics[width=0.9\columnwidth]{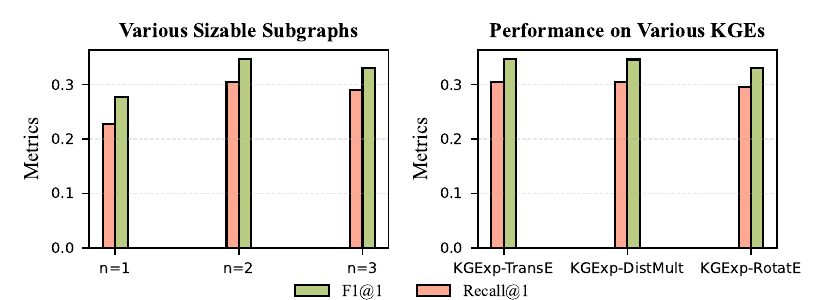}
  \caption{The explainable performance of KGExplainer with different hyper-parameters over Family-rr dataset.
  }
  \label{fig:parameter}
\end{figure}
\subsection{Hyper-parameter Sensitivity Analysis (Q3)}\label{sec:hyper_param}
In this section, to answer \textbf{Q3}, we conduct hyper-parameter sensitivity analysis on the Family-rr dataset to study the impact of different sizable explored subgraphs and various KGE-based KGC models.

\subsubsection{Size of subgraph-based explanation}
To investigate the influence of different sizable explanation subgraphs, we conduct experiments based on KGExp-RotatE by varying $n$ to \textit{1}, \textit{2}, and \textit{3}. The KGExplainer explored subgraphs with $n=1$ is equal to exploring the explainable paths.
As illustrated in the left of Figure~\ref{fig:parameter}, we observe the best performance on $F1@1$ and $Recall@1$ metrics when utilizing the explainable subgraphs with $n=2$. Meanwhile, the performance across different sizes collapsed into a hunchback shape. The reason could be that the explored subgraphs with $n=1$ are inefficient in recovering the original predictions and the subgraph patterns with $n=3$ may contain many irrelevant facts, which introduce much noise and reduce the performance of KGExplainer.

\subsubsection{KGExplainer over various KGE-based KGC models}\label{param:kges}
We investigate the impact of different target KGEs, including TransE, DistMult, and RotatE. As shown on the right side of Figure~\ref{fig:parameter}, the results indicate that KGExplainer achieves similar levels of explanation performance across $F1@1$ and $Recall@1$ for different variants. This is because KGExplainer selects key entities by perturbing neighboring entities and evaluating their influence on the prediction score, which is independent of the KGE-based KGC models. Our findings show that KGExplainer is effective in searching explanations for embedding-based models and is model-agnostic, as demonstrated by its performance across various KGEs.
\begin{figure}[t]
  \includegraphics[width=0.92\columnwidth]{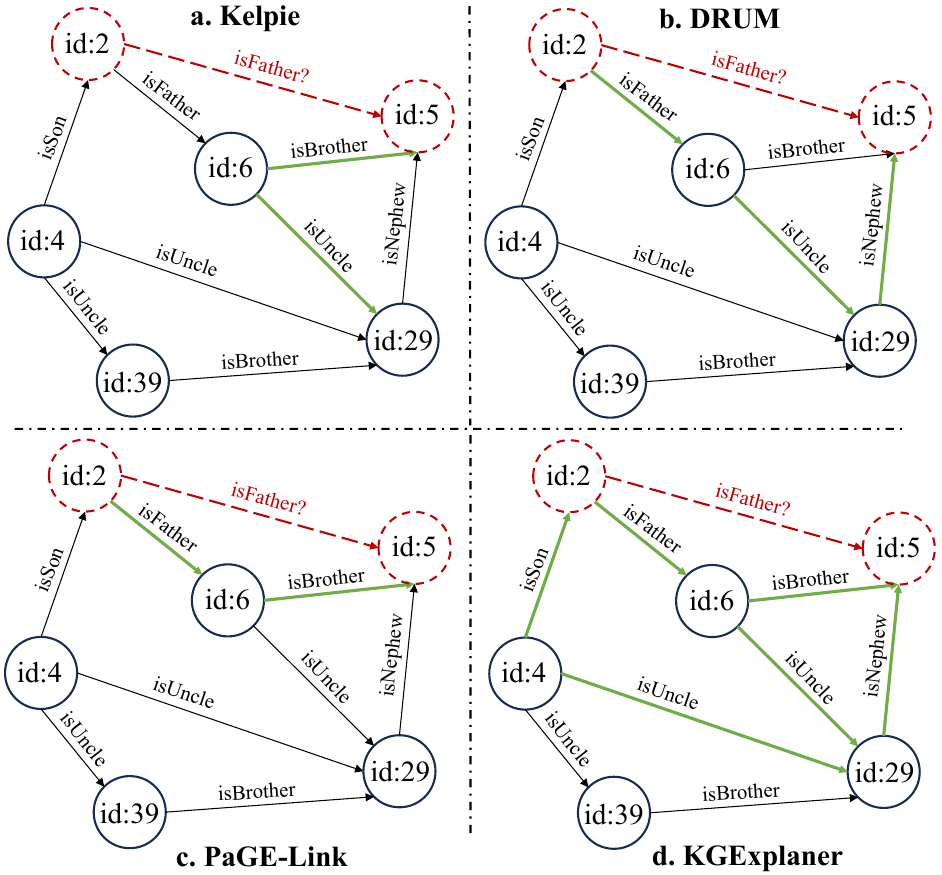}
  \caption{The explanations explored by Kelpie, DRUM, PaGE-Link, and KGExplainer on the fact $\left<id:2, isFather, id:5\right>$.
  }
  \label{fig:case_study}
\end{figure}

\subsection{Human Evaluation (Q4)} 
Following~\cite{zhang2023page}, we conduct a human evaluation by randomly picking 20 predicted links from the test set of Family-rr and generating explanations for each link using a fact-based model (i.e., Kelpie), two path-based methods (i.e., DRUM and PaGE-Link), and KGExplainer. We designed a survey with single-choice questions. In each question, we represent the predicted link and those four explanations with both the graph structure and the node/edge type information, similarly as in Figure~\ref{fig:case_study} but excluding method names. We sent the survey to people across graduate students, PhD students, and professors, including people with and without background knowledge about KGs. We ask respondents to ``please select the best explanation of \textbf{\textit{Why the model predicts this relationship between source and target nodes?}}''. At least three answers from different people are collected for each question. In total 78 evaluations were collected and 83.3\% (65/78) of them selected explanations by KGExplainer as the best. The survey demonstrates that KGExplainer is more effective in searching the human-understandable explanations than
fact- and path-based methods.

\subsection{Efficiency Analysis (Q5)}
\subsubsection{Inference Time}
We trained KGExplainer and baselines using a server with 12 virtual Intel(R) Xeon(R) Platinum 8255C CPU and one RTX 2080 Ti GPU. To verify the effectiveness of KGExplainer in searching for explanations, we designed an experiment to evaluate KGExplainer. As illustrated in Table~\ref{tab:efficent}, we show the average cost time of exploring explanations over the test sets of all datasets from different methods. We find that KGExplainer has a similar cost to CRIAGE and is more efficient than PaGE-Link and Kelpie, which indicates that KGExplainer can be effectively extended to complex KGs in searching subgraph-based explanations. This is because KGExplainer searches explanations within the enclosing subgraph and the number of calculations for each hop is only related to the average degree, which reduces the computational complexity of the exploring process. Overall, KGExplainer has superiority in exploring subgraph-based explanations with complex patterns, which can scale to large knowledge graphs.
\begin{table}[t]
\caption{The average cost time (s) of fact-based, path-based methods, and KGExplainer (i.e., KGExp-RotatE) to explore explanations on the test set of different datasets.}
\label{tab:efficent}
\begin{tabular}{lcccc} \toprule
                & \textbf{CRIAGE} & \textbf{Kelpie} & \textbf{PaGE-Link} & \textbf{Ours} \\ \midrule
\textbf{Family-rr} & \underline{1.652}             & 4.849          & 1.747          & \textbf{1.321}      \\ 
\textbf{WN-18} & \textbf{1.554}             & 4.698          & 1.744          & \underline{1.587}      \\
\textbf{FB15k-237} & \textbf{15.13}             & 28.51          & 16.93          & \underline{16.81}      \\
\bottomrule          
\end{tabular}
\end{table}
\begin{table}[]
\centering
\caption{The size and $F1@1$ performance of explanation subgraphs of different explainers on Family-rr. KGExplainer is equivalent to the path-based methods when $n=1$.}
\label{tab:sparsity}
\begin{tabular}{l|cc} \toprule
Methods                       & Avg \#edges & F1@1  \\ \midrule
Keplie (Fact-based)     & 8.297       & 20.33 \\
PaGE-Link (Path-based) & 3.085       & 21.89 \\ \midrule
KGExp-TransE ($n=1$)          & 3.182       & 28.77 \\
KGExp-TransE ($n=2$)          & 9.782       & 30.47 \\
KGExp-TransE ($n=3$)          & 15.132      & 30.43 \\ \bottomrule
\end{tabular}%
\end{table}

\subsubsection{Sparsity of Explanations}

We discuss the sparsity of the explanations from various explainers. For KGExplainer and baseline methods, we adopt the average number of edges (i.e., \textbf{Avg. \#edges}) as the size of explanations. For KGExplainer, the users can adjust the sparsity of explanation subgraphs by selecting different top-$n$ nodes per hop (Refer to Section 4.4.1). As shown in Table~\ref{tab:sparsity}, we observe that the path-based method (e.g., PaGE-Link) exploring a coherent reasoning chain with fewer facts is more efficient than the fact-based method (e.g., Keplie). Additionally, KGExplainer exploring connected subgraphs can achieve superior performance on comparable sparse explanations than the path-based method. The phenomenon in Table ~\ref{tab:sparsity} shows that KGExplainer can yield better explainability on explanations with low information density than baselines. More details can be found in Appendix~\ref{appendix:sparsity}.

\section{Conclusion and future work}
In this work, we proposed a KGExplainer to identify connected subgraphs as explanations over complex KGs and design two quantitative metrics for evaluating them. Specifically, KGExplainer used a greedy search algorithm and distilled an evaluator to find and assess key subgraphs. 
Extensive experiments demonstrate that KGExplainer is effective in providing meaningful explanations. In the future, we will generalize our framework into multiple domains, including but not limited to vision- and text-based applications.
\clearpage
\bibliographystyle{ACM-Reference-Format}
\bibliography{main}

\appendix
\section{Retraining Strategy}\label{retrain}

\begin{figure}[h]
  \includegraphics[width=1\columnwidth]{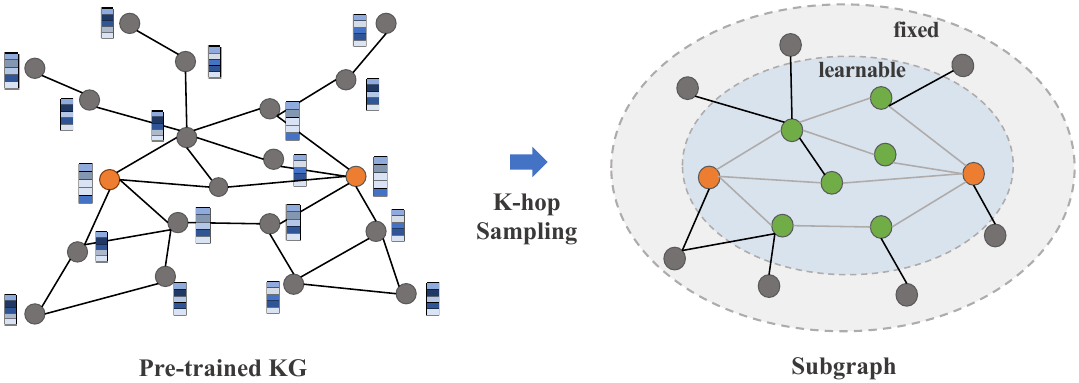}
  \caption{An example of creating a subgraph from $\mathcal{G}$. On the
left, two entities (the orange vertices) were predicted
to be connected. A subgraph was created on the right. We
first initialized $g$ with all the entities (the orange and green
vertices) within $\mathcal{G}$. Then it was expanded by including the
1-hop neighbors $\mathcal{N}_{extend}$ of existing entities (the gray vertices).}
  \label{fig:retrain}
\end{figure}

Retraing the embeddings over the entire KG $\mathcal{G}$ is infeasible due to removing one entity and its facts have little impact on the embedding of KG elements. To reduce the overhead of retraining, we developed a subgraph-based retraining strategy as shown in Figure~\ref{fig:retrain}. We extract a 2-hop enclosing subgraph $g$ around the target fact $\left<h,r,t\right>$. We argue that the subgraph $g$ contains the key information that leads to the target prediction. Thus the retraining process is only visible to the subgraph. Figure~\ref{fig:retrain} shows an example of how $g$ is extracted. Meanwhile, to ensure that the retrained entity representation space is consistent with the original knowledge graph, we expand $g$ by including the 1-hop neighbors $\mathcal{N}_{extend}$ of the existing entities and fixed their embeddings. In Figure~\ref{fig:retrain}, we show the learnable and fixed entities in the fine-tuning process. Specifically, we adopt the following constraints: (a) we use the embeddings of the entities in $\mathcal{G}$ to initialize the embeddings of $g$; (b) we fixed the relation embeddings in $g$ as the same as those in $\mathcal{G}$; and (c) we fixed the embedding of $v\in \mathcal{N}_{extend}$.
\begin{figure}[h]
  \includegraphics[width=0.95\columnwidth]{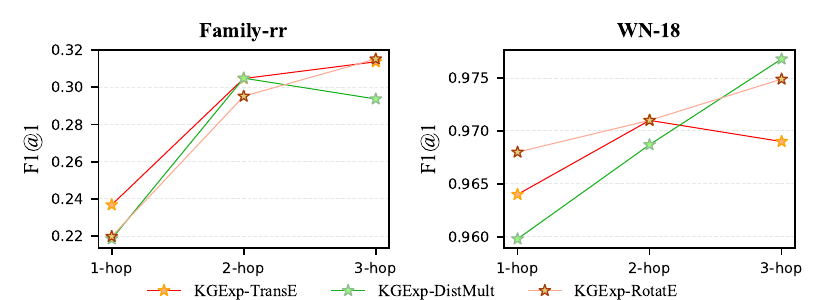}
  \caption{The $F1@1$ performance of KGExplainer on Family-rr and WN-18 over various sizable enclosing subgraphs.}
  \label{fig:khop_size}
\end{figure}

\section{Performance on Various Sizable Enclosing Subgraphs}
Similar to the experiments in Section~\ref{sec:hyper_param} (i.e., \textit{Hyper-parameter Sensitivity Analysis}), we conduct another experiment to study the model's explainability on various sizable enclosing subgraphs. We select top-$n=2$ entities per hop and vary the size of $k$-hop enclosing subgraph to $k=1,2,3$. The $F1@1$ performance of KGExplainer over various sizable subgraphs is depicted in Figure~\ref{fig:khop_size}. Theoretically, the larger the subgraph to be searched, the more sufficient key information it contains, so that more effective explanatory subgraphs can be retrieved. We can observe that KGExp-RotatE performs better as the size of subgraphs increases on both Family-rr and WN-18 datasets, which indicates that KGExplainer based on RotatE has generalization and stability. However, KGExplainer over TransE and DistMult only fits this phenomenon on one of the datasets, showing they are limited in generalizing to other data distributions. Thus to achieve a fair comparison, we consider the KGExp-RotatE as the subgraph evaluator and assess other baseline methods.

\section{Extend Experiment}\label{appdix:experiment}
\subsection{Experiment Settings}\label{appendix:setting}
\begin{table}[t]
\caption{Hyperparameters of KGExplainer in \textit{explanation phase}.}
\label{tab:exp_param}
\begin{tabular}{lccc}
\toprule
\textbf{Hyper parameter}             & \textbf{WN18} & \textbf{Family-rr} & \textbf{FB15k-237} \\ \midrule
\#entity per hop $n$        & 2             & 2                  & 2                  \\
enclosing subgraph size $k$ & 2             & 2                  & 1                  \\
target KGE                  & RotatE        & TransE             & RotatE  \\ \bottomrule          
\end{tabular}
\end{table}

\begin{table*}[t]
\centering
\caption{The $F1@N$(\%) performance of explanations over various datasets. The boldface denotes the highest score.}
\label{tab:appendx_f1}
\begin{tabular}{l|lllllllll}
\toprule
\multirow{2}{*}{\textbf{Methods}} & \multicolumn{3}{c}{Family-rr}                                                   & \multicolumn{3}{c}{WN-18}                                                       & \multicolumn{3}{c}{FB15k-237}                                                   \\ \cline{2-10}
                                  & \multicolumn{1}{c}{F1@1} & \multicolumn{1}{c}{F1@3} & \multicolumn{1}{c}{F1@10} & \multicolumn{1}{c}{F1@1} & \multicolumn{1}{c}{F1@3} & \multicolumn{1}{c}{F1@10} & \multicolumn{1}{c}{F1@1} & \multicolumn{1}{c}{F1@3} & \multicolumn{1}{c}{F1@10} \\ \midrule
CRIAGE                            & 20.75                    & 27.22                    & 36.44                     & 63.78                    & 72.91                    & 77.78                     & 40.99                    & 48.81                    & 53.26                     \\
Keplie                            & 20.33                    & 27.18                    & 36.25                     & 66.36                    & 73.58                    & 79.39                     & 36.64                    & 40.77                    & 48.01                     \\ \midrule
DRUM                              & 21.26                    & 25.23                    & 30.73                     & 71.41                    & 75.58                    & 77.80                     & 65.53                    & 67.34                    & 71.03                     \\ 
ELPE                              & 22.75                    & 29.71                    & 40.40                     & 75.87                    & 78.35                    & 83.21                     & 70.11                    & 29.71                    & 40.40                     \\
PaGE-Link                         & 21.89                    & 28.42                    & 38.49                     & 67.70                    & 70.99                    & 75.21                     & 66.12                    & 70.77                    & 76.51                     \\ \midrule
KGExp-rand                        & 19.16                    & 27.55                    & 41.48                     & 31.22                    & 37.43                    & 40.46                     & 55.17                    & 61.03                    & 66.58                     \\
KGExp-TransE                      & 30.47                    & \textbf{45.89}           & \textbf{62.96}            & \textbf{97.10}           & 98.57                    & \textbf{99.58}            & 82.04                    & 85.41                    & 89.97                     \\
KGExp-DistMult                    & \textbf{30.49}           & 44.08                    & 57.47                     & 96.87                    & 97.89                    & 99.41                     & \textbf{84.29}           & 86.74                    & 90.17                     \\
KGExp-RotatE                      & 29.52                    & 39.56                    & 57.25                     & 97.10                    & \textbf{98.58}           & 99.57                     & 83.65                    & \textbf{87.41}           &  \textbf{92.97}         \\ \bottomrule  
\end{tabular}%
\end{table*}

\begin{table*}[t]
\centering
\caption{The $Recall@N$(\%) performance of explanations over various datasets. The boldface denotes the highest score.}
\label{tab:apendx_recall}
\begin{tabular}{l|lllllllll} \toprule
\multirow{2}{*}{\textbf{Methods}} & \multicolumn{3}{c}{Family-rr}                                                               & \multicolumn{3}{c}{WN-18}                                                                   & \multicolumn{3}{c}{FB15k-237}                                                            \\ \cline{2-10}
                                  & \multicolumn{1}{c}{Recall@1} & \multicolumn{1}{c}{Recall@3} & \multicolumn{1}{c}{Recall@10} & \multicolumn{1}{c}{Recall@1} & \multicolumn{1}{c}{Recall@3} & \multicolumn{1}{c}{Recall@10} & \multicolumn{1}{c}{Recall@1} & \multicolumn{1}{c}{Recall@3} & \multicolumn{1}{c}{Recall@10} \\ \midrule
CRIAGE                            & 28.57                        & 32.02                         & 36.95                        & 86.69                        & 95.45                         & 99.34                        & 68.23                        & 76.87                         & 82.11                     \\
Keplie                            & 27.62                        & 31.87                         & 36.55                        & 88.31                        & 93.69                         & 99.51                        & 60.01                        & 66.91                         & 75.52                     \\ \midrule
DRUM                              & 16.20                        & 22.09                         & 30.92                        & 91.12                        & 93.69                         & 94.93                        & 87.88                        & 89.01                         & 91.12                     \\
ELPE                              & 29.84                        & 35.17                         & 41.34                        & 89.97                        & 93.11                         & 95.23                        & 85.58                        & 89.76                         & 93.21                     \\
PaGE-Link                         & 27.14                        & 31.77                         & 38.13                        & 85.36                        & 88.32                         & 92.20                        & 85.13                        & 87.32                         & 89.04                     \\ \midrule
KGExp-rand                        & 26.98                        & 33.44                         & 43.97                        & 86.09                        & 92.47                         & 94.61                        & 50.37                        & 57.12                         & 64.31                     \\
KGExp-TransE                      & \textbf{34.61}               & \textbf{46.85}                & \textbf{61.61}               & \textbf{98.79}               & \textbf{99.47}                & \textbf{99.84}               & \textbf{92.07}               & \textbf{93.41}                & 94.57                     \\
KGExp-DistMult                    & 34.60                        & 43.05                         & 56.26                        & 98.14                        & 99.31                         & 99.71                        & 91.67                        & 92.01                         & 93.89                     \\
KGExp-RotatE                      & 33.02                        & 41.48                         & 56.26                        & 98.75                        & 99.47                         & 99.84                        & 91.45                        & 93.01                         & \textbf{94.69}       \\ \bottomrule    
\end{tabular}%

\end{table*}
\begin{table}[t]
\caption{Hyperparameters of KGExplainer in \textit{distillation phase}.}
\label{tab:dis_param}
\begin{tabular}{lccc}
\toprule
\textbf{Hyper parameter} & \textbf{WN18} & \textbf{Family-rr} & \textbf{FB15k-237} \\ \midrule
embedding dim            & 64            & 64                 & 64                 \\
learning rate            & 0.001         & 0.0015             & 0.001              \\
weight of pair-wis loss   & 0.5           & 0.5                & 0.5                \\
epochs                   & 20            & 20                 & 20                 \\
R-GAT layers             & 3             & 2                  & 3                  \\
optimizer                & Adam          & Adam               & Adam               \\
batch size               & 2048          & 1024               & 4096 \\
\bottomrule
\end{tabular}
\end{table}
\subsubsection{Settings of KGExplainer}
The experiments have two phases: (1) Generating subgraph-based explanations; and (2) Distilling the subgraph evaluator for inference. The details of hyper-parameter settings are reported in Table~\ref{tab:exp_param} and Table~\ref{tab:dis_param}.

In the \textit{explanation phase}, we mainly focus on the size of enclosing subgraphs, the complexity of explanation subgraphs, and the target KGE models. For different datasets, we set various parameters and published them in Table~\ref{tab:exp_param}.
In the \textit{distilling phase}, we are mainly concerned with the efficiency of training or inference and the prediction performance of the distilled subgraph evaluator on the KGC task. Specifically, we try different parameters over various datasets on model architecture and optimization for distilling an effective evaluator. The details of the parameter settings are shown in Table~\ref{tab:dis_param} 

\subsubsection{Implementation details}
The experiments have two phases: (1) Generating subgraph-based explanations; and (2) Distilling the subgraph evaluator for inference.

In the \textit{explanation phase}, to efficiently search for informative facts, we extract key patterns from the 2-hop enclosing subgraph surrounding the target prediction. We set the parameter $n=2$ to determine the number of selected entities per hop during greedy searching. To achieve the best evaluation capability, we use RotatE~\cite{sun2019rotate} as the target KGE and distill an evaluator to assess explainability.
In the \textit{distilling phase}, we adopt an R-GAT with three layers for all datasets and KGE models.
To ensure sufficient expressive power for the model to distill,
the hidden dimension is set to 64 for both pre-trained KG embeddings and subgraph representations of the distilled model. To balance the alignment and ranking objective, we set the weight of ranking loss to $\lambda=0.5$. During training for all datasets, we use a learning rate of 0.001. We model explanation methods based on the distilled model from RotatE, to ensure a fair and effective evaluation of explainability.
\begin{table*}[t]
\caption{The sparsity and corresponding performance of different explainers.}
\label{tab:appendix_sparsity}
\begin{tabular}{l|ll|ll|ll} \toprule
\multicolumn{1}{c}{}                                   & \multicolumn{2}{c}{\textbf{Family-rr}}           & \multicolumn{2}{c}{\textbf{WN-18}}               & \multicolumn{2}{c}{\textbf{FB15k-237}}           \\ 
\multicolumn{1}{c}{\multirow{-2}{*}{\textbf{Methods}}} & Avg. \#edges & {F1@1} & Avg. \#edges & {F1@1} & Avg. \#edges & {F1@1} \\ \midrule
Keplie (fact-based)                                                 & 8.297        & 20.33                             & 4.794        & 66.37                             & 11.190        & 36.64                             \\
PaGE-Link (path-based)                                              & 3.085        & 21.89                             & 3.640         & 67.70                              & 4.993        & 66.12                             \\ \midrule
KGExp-TransE ($n=1$)                                     & 3.182        & 28.77                             & 2.263        & 96.40                             & 5.142        & 75.67                             \\
KGExp-TransE ($n=2$)                                     & 9.782        & 30.47                             & 4.772        & 97.10                              & 13.661       & 82.04                             \\
KGExp-TransE ($n=3$)                                     & 15.132       & 30.49                             & 9.223        & 96.91                             & 26.146       & 83.21                            \\ \bottomrule
\end{tabular}
\end{table*}
\subsubsection{Details of baselines}
We implement our KGExplainer and baseline models in Pytorch~\cite{paszke2017automatic} and DGL~\cite{wang2019dgl} library on an RTX 2080Ti GPU with 11GB memory. For the implementation of baselines, we use the source code of Kelpie~\cite{rossi2022explaining} to reproduce the faithful results and explore explanations of fact-based methods CRIAGE~\cite{pezeshkpour2019investigating} and Kelpie. For other baselines, we adopt the source code reported by their original paper and use their pre-defined parameters to produce explanations. In addition to comparing against baselines with no parameter settings for certain datasets, we also examine the parameters used in a similar scale dataset. We report the average results and their standard deviation by five runs in Section~\ref{sec:experiment}.

\subsection{The Full Results of Explanations}\label{appendix:full_results}
The full results (i.e., $F1@[1,3,10]$ and $Recall@[1,3,10]$) of the explanations are shown in Table~\ref{tab:appendx_f1} and Table~\ref{tab:apendx_recall}. We can observe that KGExplainer is also the best.

\subsection{Sparisity of Explainers}\label{appendix:sparsity}

We discuss the sparsity of the explanations from various explainers.
For Fact- and Path-based methods, we directly adopt the edges and paths for further evaluation. Therefore the size of explanations depends on the number of edges and path lengths they explore. For KGExplainer, the users can adjust the sparsity of explanation subgraphs by selecting different top-$n$ nodes per hop (See Section 4.4.1). Once the model is trained, the user can directly adjust the value of $n$ without retraining. For various explainers, We present the average number of edges (i.e., \textbf{Avg. \#edges}) and corresponding $F1@1$ performance on all datasets in Table~\ref{tab:appendix_sparsity}. We see that the path-based method (e.g., PaGE-Link) with a complete reasoning chain is more effective than the fact-based method (e.g., Keplie) on fewer visible facts. Moreover, we observe that KGExplainer can dynamically adjust the size of the explanation subgraphs and achieve superior performance compared with Fact- and Path-based methods.

\section{Discussion}
\subsection{Discussion with Related Works}
In this section, we further discuss the novelty of KGExplainer compared with recent related methods.
The works~\cite{han2020explainable,joshi2020searching} mainly focus on pruning subgraphs to eliminate irrelevant facts and to improve the performance of link prediction on either temporal or static knowledge graphs. To some extent, the pruned subgraphs can be considered key information for predicted links. However, the pruned subgraphs are not connected but rather a collection of facts, which are referred to as Fact-based explanation methods. 
On the other hand, the paper~\cite{lin2018fact} uses rule-based methods to detect fixed patterns and check specific facts, which is different from KG-based explanation methods for knowledge graph completion tasks. Similarly, the methods presented in~\cite{zhao2023ke,baltatzis2023kgex} are post-hoc explanation methods that identify key triples as explanations. The identified key triples are discrete facts and can not form a reasoning chain. These two methods can be classified as \textbf{Fact-based} models. For the KGC task, the connected subgraphs can form a reasoning chain, which can better explain target prediction~\cite{zhang2023page}. We proposed KGExplainer to focus on exploring connected subgraphs are different from previous models. In summary, our method can contribute to the knowledge graph reasoning community and promote the KG-based explanation methods.

\subsection{Discussion of Distilled Results} \label{dis:disti}
In Table~\ref{tab:faith}, we can see that KGExplainer benefits the base model for some datasets.
We observe the same behavior in~\cite{deng2020can,jiao2019tinybert}.
From their insights, we think the KGExplainer distilled from KGEs may remove some noise and irrelevant information to the KGC task, making KGExplainer more robust and having better generalization ability, which leads to KGExplainer performing better than base KGEs to some extent.








\end{document}